\pdfoutput=1

\documentclass[11pt]{article}

\usepackage[final]{acl}

\usepackage{times}
\usepackage{latexsym}

\usepackage[T1]{fontenc}

\usepackage[utf8]{inputenc}

\usepackage{microtype}

\usepackage{inconsolata}

\usepackage{graphicx}

\usepackage{amsmath}
\usepackage{amssymb}

\usepackage{booktabs}
\usepackage{wrapfig}
\usepackage{multicol}
\usepackage{multirow}
\usepackage{graphicx}
\usepackage{float}
\usepackage{arydshln}
\usepackage[linesnumbered,ruled,vlined]{algorithm2e}

\newcommand{\MODELNAME}{RPO}

\title{RPO: Retrieval Preference Optimization for Robust Retrieval-Augmented Generation}

\author{
Shi-Qi Yan\textsuperscript{1}, Quan Liu\textsuperscript{2} \and Zhen-Hua Ling\textsuperscript{1}  \\
  \textsuperscript{1}National Engineering Research Center of Speech and Language Information Processing, \\
      University of Science and Technology of China \\
  \textsuperscript{2}State Key Laboratory of Cognitive Intelligence, iFLYTEK Research \\
{\tt sqyan01@mail.ustc.edu.cn},  {\tt quanliu@iflytek.com},  {\tt zhling@ustc.edu.cn}
}

\begin{document}
\maketitle
\begin{abstract}
While Retrieval-Augmented Generation (RAG) has exhibited promise in utilizing external knowledge, its generation process heavily depends on the quality and accuracy of the retrieved context.
Large language models (LLMs) struggle to evaluate the correctness of non-parametric knowledge retrieved externally when it differs from internal memorization, leading to \emph{knowledge conflicts} during response generation.
To this end, we introduce the \textbf{R}etrieval \textbf{P}reference \textbf{O}ptimization (\MODELNAME{}), a lightweight and effective alignment method to adaptively leverage multi-source knowledge based on retrieval relevance.
An implicit representation of retrieval relevance is derived and incorporated into the reward model to integrate retrieval evaluation and response generation into a single model, solving the problem that previous methods necessitate the additional procedure to assess the retrieval quality.
Notably, \MODELNAME{} is a RAG-dedicated alignment approach that quantifies the awareness of retrieval relevance in training, first overcoming mathematical obstacles.
Experiments on four datasets demonstrate that \MODELNAME{} outperforms RAG by 4-10\% in accuracy without any extra component, exhibiting its robust generalization.
\end{abstract}
\section{Introduction}

\begin{figure}[t]
  \centering
  \includegraphics[width=0.5\textwidth]{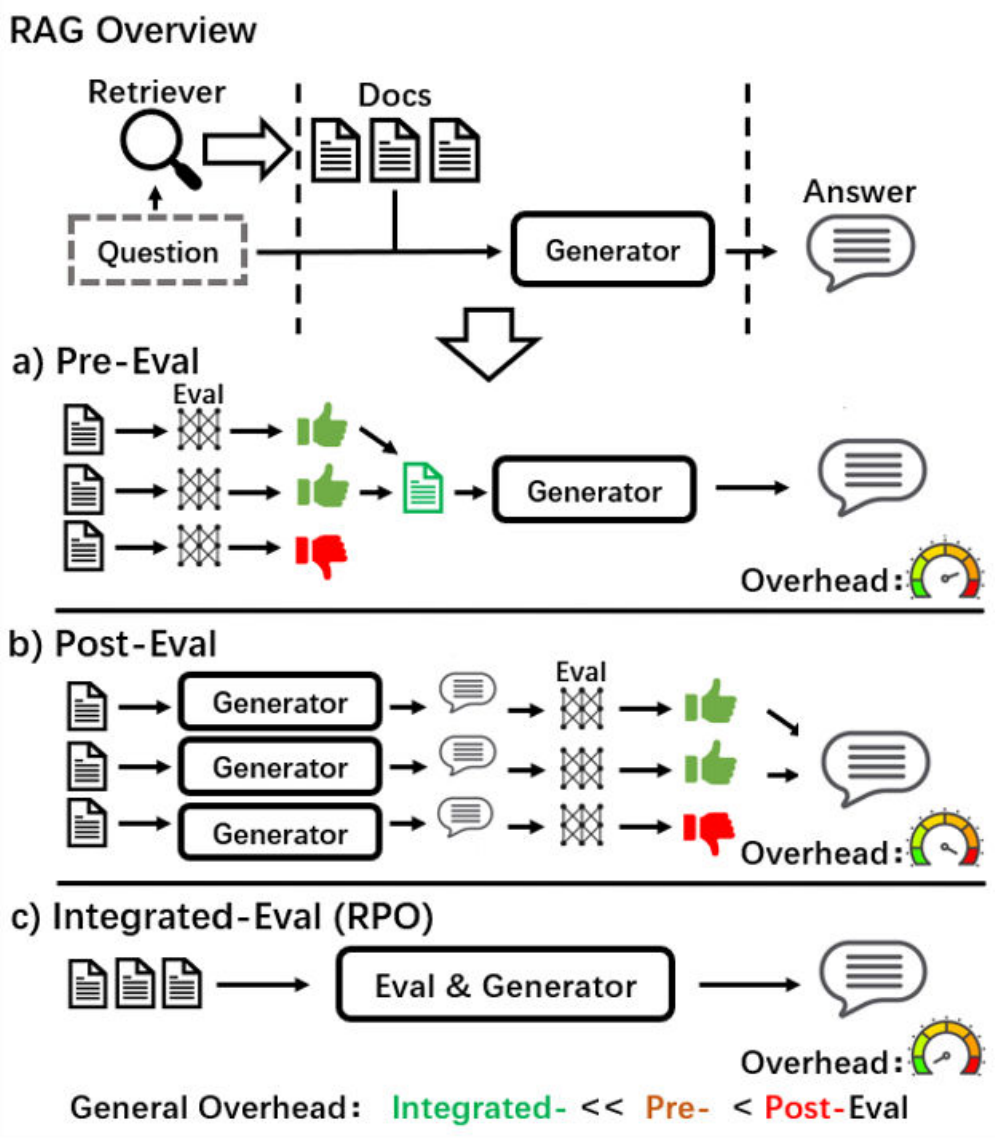}
  \caption{
  The figure showcases the overview of RAG and three categories of adaptive RAG, including a) Pre-Eval, b) Post-Eval, and c) Integrated-Eval approaches. The estimated computational overhead of three categories is demonstrated as well, exhibiting the efficiency of our \MODELNAME{} in inference.
  }
  \label{fig-demo}
\end{figure}

Despite the wide application in natural language processing tasks, large language models (LLMs) still struggle with knowledge-intensive tasks~\citep{ralm,knowledge-intensive}.
As a general and effective approach, retrieval-augmented generation (RAG)~\citep{rag,rag-2021} involves retrieving the context related to the input query from an external corpus and integrating it for generation.

However, RAG has been found to have the potential for over-reliance on retrieval, which could unconsciously lead to hallucination, particularly when the information retrieved, also called non-parametric knowledge, conflicts with the parametric knowledge embedded within LLMs~\citep{knowledge_conflict-2021,knowledge_conflict-survey}.
Specifically, RAG tends to prioritize the retrieved external context over the internal knowledge when conflicts arise~\cite{PoisonedRAG,robustrag,crag}.
Therefore, the performance of RAG depends heavily on the accuracy of the retrieval process, as inaccurate retrievals can introduce irrelevant or even harmful information, affecting the quality of generated text~\citep{irrelevant-ctx,rony2022dialokg}.
To address the challenge, previous studies evaluated the quality of retrieval before (pre-eval) or after generation (post-eval).
However, as shown in figure~\ref{fig-demo}, such approaches called adaptive RAG require extra processing to evaluate the value of retrieval via several API or LLM calls, leading to massive computational overhead.
Meanwhile, removing part of the negative context that is assessed by the evaluator reduces the information provided for generation.
It makes the generator more dependent on the evaluator, affecting the ultimate performance as well.

Considering the issues above, in this paper, we propose \MODELNAME{}, a \textbf{R}etrieval \textbf{P}reference \textbf{O}ptimization algorithm,
aiming to enhance the robustness of LLM to multi-source knowledge by integrating retrieval evaluation in generation through reinforcement learning.
A comprehensive theoretical analysis is first conducted to highlight the technical limitations of previous preference optimization algorithms~\citep{rlhf,dpo,knowpo} in the context of the RAG scenario.
We mathematically prove the limitations of the previous methods, which violate the objective of adaptive RAG, which is to select the correct answer both before and after retrieval.
When conflict is involved between parametric and non-parametric knowledge, an over-tendency towards the retrieved knowledge still easily arises during the generation.
Building on this theory, our \MODELNAME{} alignment method is designed to mitigate over-reliance on retrieval by incorporating the awareness of retrieval relevance into the reward model.
To strengthen the capability of conflict mitigation, \MODELNAME{} simulates knowledge conflict and rectifies the discernment of LLM about which type of knowledge to prioritize.
First, we instructed LLM to generate answers with and without retrieval respectively, filtering the contradictory instances as knowledge conflict.
In the meantime, the relevance of the retrieved context is quantified and represented implicitly.
Ultimately, the calculated relevance is integrated into the reward model for alignment to adaptively reward the positive answer in the contradictory pair based on the quality of retrieval.

As shown in figure~\ref{fig-demo}, \MODELNAME{} (Integrated-eval) integrated the evaluation of the retrieval quality with the generation, without any additional overhead, exhibiting significant efficiency.
Meanwhile, results on four datasets of PopQA~\citep{popqa}, Natural Questions~\citep{nq}, TriviaQA~\citep{triviaqa}, and RGB~\citep{rgb} show that \MODELNAME{} can significantly improve the performance of RAG over prior approaches, demonstrating its consistent advancements across various benchmarks. 

In summary, our contributions in this paper are three-fold:
1) We propose an optimization strategy named \MODELNAME{}, aimed at encouraging LLMs to synchronously evaluate the retrieved context and selectively leverage non-parametric knowledge without any explicit processing during response generation.
2) We provide a mathematical proof highlighting the inadequacy of existing preference optimization strategies for direct application in RAG-based scenarios and propose a more efficient algorithm as well as a data collection method for training to address this limitation.
3) Through experimentation involving multiple LLMs and benchmarks, we validate the efficacy of our proposed \MODELNAME{} algorithm and showcase its consistent performance advancements.

\section{Related Work}

\paragraph{Adaptive RAG}
In traditional RAG~\citep{rag} applications, the retrieved context, referred to as non-parametric knowledge, may sometimes conflict with the parametric knowledge stored in LLMs.
Previous research has explored the evaluation of retrieval quality and the adaptive use of non-parametric knowledge for conflict resolution, which can be generally categorized into pre-eval and post-eval approaches.
Pre-eval methods\citep{retrobust,crag,astuterag} involve employing a specialized classification language model (LM) or instructing LLMs to assess retrieval quality.
In contrast, post-eval methods\citep{self-rag,robustrag} entail independently generating multiple responses based on various retrieved documents and selecting the best answer as the final response.
However, on the one hand, both approaches are computationally demanding and structurally complex, resulting in decreased inference efficiency.
On the other hand, part of the information is removed by the evaluator, making the generator more dependent on the performance of the evaluator, which affects the ultimate performance as well.

\paragraph{Model Alignment}
In reviewing the Reinforcement Learning from Human Feedback (RLHF)~\citep{rlhf} pipeline, three main phases are included: supervised fine-tuning (SFT), reward model learning, and RL optimization.
After fine-tuning a pre-trained LM a pair of answers is sampled $(y_1, y_2)\sim \pi_\text{SFT}(y\mid x)$, crowd workers annotate the preferred one between the pair, denoted as $y_w \succ y_l \mid x$.
A latent reward model is introduced and learned afterward to quantify the preference.
Ultimately, the Proximal Policy Optimization (PPO)~\citep{ppo} algorithm is adopted as the objective of RL optimization.
Afterward, as one of the most popular alignment strategies, DPO~\citep{dpo} involves replacing the external reward model with a closed-form expression.
Instead of learning an explicit reward model, DPO reparameterizes the reward function $r$ using a closed-form expression with the optimal policy.
The computationally lightweight approach significantly eliminates the need for direct RL optimization and outperforms existing methods.

\section{Task Definition}


\subsection{RAG Formulation} 
To answer a question $x$ from a dataset $\mathbf{D}$ with an LLM $\pi$, RAG requires the retrieved context $R$ as the supplementary material before response generation.
In most situations, the first stage of the system is to retrieve multiple relevant documents $D^r = \{D_1^r, ... D_K^r\}$ from an accessible corpus $\mathbb{C}$, which then serve as supplementary input to the query for the LLM generation.  
Thus the RAG task can be simplified into:
\begin{equation}
y_{n+p} = \pi(x, R) |_{R = D^r},
\label{equ-form_RAG}
\vspace{-1mm}
\end{equation}
where $y_{n+p}$ means the answer for the question $x$ that has access to the retrieved results, i.e., all retrieved context $D^r$. LLMs autonomously select either parametric or non-parametric knowledge for response generation.

\subsection{Knowledge Conflict} \label{sec-conflict}
Apart from the response that integrates retrieved information, $\pi$ actually has its own potential answer with the knowledge memorized in the parameters.
It can be activated by directly instructing $\pi$ to generate the answer, expressed as:
\begin{equation}
y_p = \pi(x, R) |_{R = \phi} ,
\label{equ-form_pure_LLM}
\end{equation}
where $y_p$ means the answer without any retrieved context, i.e. null set in the equation above, representing the response with parametric knowledge for $x$.
Note that if the parametric knowledge and retrieved non-parametric knowledge are different, i.e., knowledge conflict arises, the generator in RAG should make a decision on which knowledge to be referred to.
If the knowledge from the retrieved context is adopted, the answer would be vary from $y_p$.
Based on this situation, we filtered the non-parametric answers $y_n$ from $y_{n+p}$ 
Ultimately, we can detect knowledge conflict and filter non-parametric answers by:
\begin{equation}
\text{Acc}(y_n) + \text{Acc}(y_p) = 1,
\label{equ-KC_def}
\end{equation}
where $y_n \in y_{n+p}$, and the correct answer can be formulated as $\text{Acc}(y) = 1$, and the incorrect one satisfies $\text{Acc}(y) = 0$.
Therefore, Equ.~(\ref{equ-KC_def}) indicates that only one in the pair of the answers is correct.

\section{Why DPO is Limited to Apply to RAG} \label{sec-preliminary}
DPO~\citep{dpo} has shown its great performance in fine-grain optimization by aligning LLMs with the chosen ones in the preference pairs, which just meets the task requirement of the knowledge conflict.
However, several concerns exist regarding the application of DPO to RAG-based tasks.

\paragraph{Firstly, the optimization objective of RLHF and DPO is inconsistent with the conflict-mitigating target in RAG.}
Considering the integrated retrieved context in the input when applied to RAG, the ultimate optimization objective of PPO-based methods such as RLHF and DPO can be formulated as :
\begin{flalign}
\begin{split}
\max_{\pi_\theta}&{\mathbb{E}_{x\sim \mathbf{D},  y\sim\pi_\theta(y\mid x)}r (x,D^r,y)} \\
 &\quad\quad-\beta\mathbb{D}_{\text{KL}}[\pi _\theta (y\mid x,D^r) \left |  \right | \pi _{\text{ref}}(y\mid x,D^r) ] ,
\label{equ-RLHF_RL_obj}
\end{split}
\end{flalign}
where $\beta$ is the controlling hyper-parameter. $\pi_{\theta}$ and $\pi_{\text{ref}}$ indicate the trainable and reference policies respectively, which are both initialized to $\pi_\text{SFT}$, while $\pi_{\text{ref}}$ is frozen.
The last term in the formulation is adopted as an extra constraint, which is significant in preventing the model from deviating to far from the original distribution.
However, in the RAG application, LLMs require considerable parameter tuning to improve the distribution from the over-tendency on retrieved context. 
For instance, if the parametric answer is the preferred one, the ideal distribution should be aligned with $\pi_{\text{ref}}(y\mid x)$, while the non-parametric answer is preferred, the target distribution should be aligned with $\pi_{\text{ref}}(y\mid x,D^r)$.
The constraint in the previous optimization strategies will affect the efficiency and the performance of the training methods, remaining bias on the non-parametric answers.

\paragraph{Secondly, the partition function within the reward model can not be canceled out.}
Note that DPO necessitates both positive and negative responses to have high probabilities for the same input, i.e., $(y_w, y_l)\sim\pi_{\text{SFT}}(y\mid x)$, satisfying that $\log\pi_{\text{SFT}}(y_w\mid x), \log\pi_{\text{SFT}}(y_l\mid x)> \epsilon$, where $\epsilon$ is a rather high value among the output log-probabilities of the policy.
When DPO is directly applied to RAG, considering the existence of the retrieval $D^r$, the expression of the DPO optimizing objective can be formulated as:
\begin{flalign}
\begin{split}
&\mathcal{L}_{DPO}(\pi_\theta;\pi_{\text{ref}}) = -\mathbb{E}_{(x,y_w,y_l)\sim \mathbf{D}}\bigg[\log\\
&{\sigma}\bigg( \beta\log{\frac{\pi_\theta(y_w|x_w)}{\pi_{\text{ref}}(y_w|x_w)}} -\beta\log{\frac{\pi_\theta(y_l|x_l)}{\pi_{\text{ref}}(y_l|x_l)}}\bigg) \\
&\quad\quad\ \pm \bigg( \beta \log{Z(x)} - \beta \log{Z(x,D^r)} \bigg) \bigg],
\label{equ-DPO_for_RAG_Loss}
\end{split}
\end{flalign}
where $x_w=x$ and the last term is positive when the parametric answer is the positive one, while $x_w=\{x, D^r\}$ and the last term is negative when the answer with non-parametric knowledge is positive. 
Detailed proof can be found in Appendix~\ref{appendix-DPO_for_RAG_Loss}.
Apparently, this loss function becomes complex and impractical to calculate due to the existence of the partition function.

\paragraph{Thirdly, over-tendency towards non-parametric knowledge is still inevitable since parametric answers are fabricated for training.}
Due to the issue of the partition function, the input of $y_n$ and $y_p$ should be the same, which does not conform to the real-world application.
Prior studies have attempted to fabricate the parametric answer and pretending that it is generated with retrieved context, i.e., $(y_n, y_p)\sim\pi_{\text{SFT}}(y\mid x, D^r)$ \citep{knowpo}.
However, the potentially significant discrepancy in likelihood between fabricated and original answers could hinder LLM convergence during training, leading to suboptimal outcomes.
For instance, the situation in the inference stage widely exists where an instance satisfies $(x_{\text{inf}}, y_{p\_\text{inf}}\succ y_{n\_\text{inf}})$ but the optimized LLM still chooses the suboptimal non-parametric answer as the final response:
\begin{flalign}
\begin{split}
\pi_{\text{DPO}}(y_w\mid x_{\text{inf}}, D^r) <\pi_{\text{DPO}}(y_l\mid x_{\text{inf}}, D^r).
\label{equ-sub_solution}
\end{split}
\end{flalign}
Equ.~(\ref{equ-sub_solution}) suggests that despite DPO is conducted for training, the optimized policy still tends to take the dispreferred answer as the response as long as a considerable discrepancy exists between the initial preferred and dispreferred answers.
Detailed proof can be found in Appendix~\ref{appendix-sub_solution}.

\section{Methodology}

\begin{figure*}[t]
  \centering
  \includegraphics[width=0.85\textwidth]{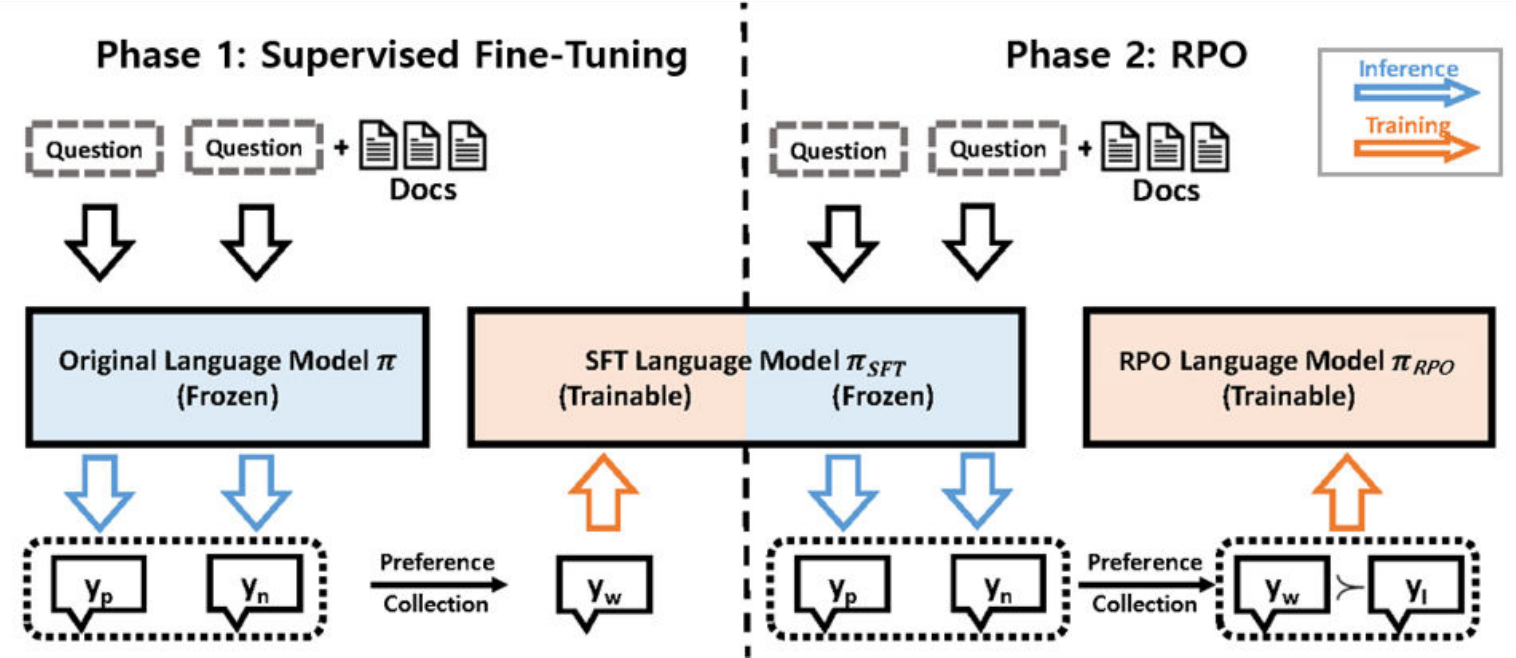}
  \caption{
  An overview of \MODELNAME{} at training. 
  In phase 1, given a question and the retrieved documents, two answers $(y_p, y_n)$ are generated by the frozen language model $\pi$. After comparing with the golden answers, instances that involve knowledge conflict are filtered for supervised fine-tuning.
  In phase two, the fine-tuned LLM is prompted to generate a pair of answers again, and the instances with knowledge conflict are filtered as the training set of \MODELNAME{}.
  }
  \label{fig-method}
\end{figure*}

Motivated by the challenges encountered in implementing preference optimization to RAG as illustrated above, this study aims to propose a RAG-specific approach for policy optimization.
Acknowledging the discrepancy between the reinforcement learning objective of the DPO and the requirements of RAG, we first propose a new reinforcement learning objective by incorporating a representation of retrieval relevance to adaptively reward LLM based on retrieval quality.
Furthermore, we outline a data collection and filtering strategy to simulate the knowledge conflict for the practical training.

\subsection{Theoretically Analysis}
\paragraph{Reward Model}
\label{sec-rm}
Since the reinforcement learning objective formulated as Equ.~(\ref{equ-RLHF_RL_obj}) has shown a discrepancy against the target of conflict mitigation in RAG, modifying the RL objective representation is primary and significant.
In this paper, we mainly attribute the discrepancy to the absence of the retrieval rewarding. 
Previous studies conventionally regard retrieved context as a fixed part of the input to build the reward model, i.e., $(y_w;y_l) \mid x, R $.
However, from the perspective of the entire RAG system, the retrieved context is only an intermediate variable, conditioned on the input query, which is consistent between preferred and dispreferred samples. 
Therefore, we suppose that the reward model in RAG should reward not only a preferred answer, but also a preferred retrieval, i.e., $(y_w,R_w;y_l,R_l)\mid x$.
Ultimately, the RL objective can be formulated as:
\begin{flalign}
\begin{split}
\max_{\pi_\theta}&{\mathbb{E}_{x\sim \mathbf{D},y\sim \pi_\theta (y\mid x,R)}r (x,y,R)} \\
 &-  \beta\mathbb{D}_{\text{KL}}[\pi _\theta (y,R\mid x) \left |  \right | \pi _{\text{ref}}(y,R\mid x) ] .
\label{equ-RaPO_RL_obj}
\end{split}
\end{flalign}

Similar to the derivation of the reward model in the DPO strategy, we can get the reward model formulation in our \MODELNAME{}:

\begin{flalign}
\begin{split}
r(x,y,R) =&\beta\log{\frac{\pi(y\mid x,R)}{\pi_{\text{ref}}(y\mid x,R)}}\\
& + \beta\log{\frac{\pi(R\mid x)}{\pi_{\text{ref}}(R\mid x)}}+ \beta\log{Y(x)},\\
\end{split}
\label{equ-rm}
\end{flalign}
where $Y(x)$ is the partition function, the details about the reward model can be found in Appendix~\ref{appendix-rapo_rm}.

\paragraph{Length Normalization}
Previous studies have observed the tendency of LLMs to be influenced by the length bias during DPO. 
In \MODELNAME{}, since retrieved context is generally much longer than the response, the length of the retrieved context could greatly affect the reward model, raising the length bias
To mitigate the excessive impact of the retrieval-awareness term, and overcome the length bias of LLMs, we utilized the average log probabilities as a part of the reward. 
Substituting the length normalization in the reward model representation, the ultimate \MODELNAME{} training objective can be written as:
\begin{flalign}
\begin{split}
&\mathcal{L}_{\text{\MODELNAME{}}} = -\mathbb{E} \bigg[ \log\sigma \bigg( \underbrace{\beta\log{\frac{\pi_\theta(y_w|x,D^r)}{\pi_{\text{ref}}(y_w|x,D^r)}}}_{\text{(a)preferred generation reward}} \\
&\underbrace{-\beta\log{\frac{\pi_\theta(y_l|x,D^r)}{\pi_{\text{ref}}(y_l|x,D^r)}}}_{\text{(b)dispreferred generation reward}}\underbrace{ {\pm\frac{\beta}{\left | D^r \right | }\log{\frac{\pi_\theta(D^r\mid x)}{\pi_{\text{ref}}(D^r\mid x)} }}}_{\text{(c)retreival reward}} \bigg) \bigg],
\label{equ-RaPO_Loss}
\end{split}
\end{flalign}
where the first and second terms (Equ.~(\ref{equ-RaPO_Loss}a), (\ref{equ-RaPO_Loss}b)) represent the preferred and dispreferred reward of generation respectively, which is consistent with DPO.
While the last term (Equ.~(\ref{equ-RaPO_Loss}c)) indicates the reward of the retrieved context, which is positive when the non-parametric answer $y_n$ is preferred against the parametric answer $y_p$, i.e., $y_n \succ y_p$, and negative when the parametric answer is preferred, i.e., $y_p \succ y_n$.

\begin{algorithm*}[t]
    \SetAlgoLined 
    \SetAlgoNoLine
    \caption{\MODELNAME{} Training Procedure}
    \SetKwInOut{Model}{Model}
    \SetKwInOut{Dataset}{Dataset($\mathbf{D}$)}
    \SetKwInOut{Output}{Output}
    \Model{$\pi$}
    \Dataset{$\mathcal{X}$ (Input Questions), $\mathcal{Y}$ (Output Labels), $\mathbb{C} = \{D_1, D_2, ..., D_N\}$ (Documents)}
    \Output{$\pi_{\text{\MODELNAME{}}}$ (Optimized Policy)}
    \tcp{Supervised Fine-Tuning} \label{cmt} \leavevmode
    \ForEach{$(x,y) \in (\mathcal{X}, \mathcal{Y})$}{
    $y_p = \pi(x)$ \\
    $y_{n+p} = \pi(x,D^r)$, $D^r=\{D^r_j, j=1,2,...K\}=$ Retriever($x$) \\
    }
    $\mathbf{D}_\text{SFT}=$Conflict\_Collection($\mathbf{D}$, Condition:$\text{Acc}(y_{n+p}) + \text{Acc}(y_p) = 1$ )\\
    $\pi_{\text{SFT}}=$Supervised\_FineTuning($\pi,\mathbf{D}_\text{SFT}$)\\
    \tcp{Retrieval Preference Optimization} \label{cmt} \leavevmode
    \ForEach{$(x,y) \in (\mathcal{X}, \mathcal{Y})$}{
    $y_p = \pi_{\text{SFT}}(x)$ \\
    $y_{n+p} = \pi_{\text{SFT}}(x,D^r)$, $D^r=\{D^r_j, j=1,2,...K\}=$ Retriever($x$) \\
    }
    $\mathbf{D}_\text{\MODELNAME{}}=$Conflict\_Collection($\mathbf{D}$, Condition:$\text{Acc}(y_{n+p}) + \text{Acc}(y_p) = 1$)\\
    $\pi_{\text{\MODELNAME{}}}=$\MODELNAME{}($\pi_{\text{SFT}},\mathbf{D}_\text{\MODELNAME{}}$)
    \label{algorithm-main}
\end{algorithm*}

\subsection{Training Overview}
In this section, we illustrate how to collect, filter, and formulate data for SFT and preference optimization.
Figure~\ref{fig-method} and Algorithm 1 present an overview of \MODELNAME{} at training.
Each example is comprised of a query and a corresponding Wikipedia page that can answer the question and has one or more short spans from the annotated passage containing the actual answer.

\paragraph{Preference Pairs Collection} \label{sec-data_collection}
We first construct the preference pairs adopted for supervised fine-tuning (SFT) and \MODELNAME{}, aimed at enhancing the model's awareness to leverage retrieved non-parametric knowledge adaptively.
Given an instance from the dataset $(x,y) \in \mathbf{D}$, we respectively instruct the model to generate responses with and without retrieval $(y_{n+p},y_p)$ as illustrated in Section~\ref{sec-conflict}.
Two subsets sampled from $\mathbf{D}$ are constructed to collect preference pairs.
In the first subset $\mathbf{D}^1$, our goal is to continually enhance the model's ability to read and comprehend the retrieved context. 
Instances are sampled where the model fails to answer the questions directly, while correctly generating the responses with retrieval, i.e., $\text{Acc}(y_{n+p})  > \text{Acc}(y_p)$.
To further confirm that $y_{n+p}$ refers to the retrieved knowledge, i.e. $y_{n+p}=y_n$, we solely select samples where the ground truths are contained in the retrieved context.
The second subset $\mathbf{D}^2$ focuses on mitigating the over-reliance of the model on the retrieved knowledge.
We select the instances where the model could have responded correctly while being affected by the retrieved knowledge and generating incorrect answers, i.e., $\text{Acc}(y_{p})  > \text{Acc}(y_{n+p})$.
Note that interference due to incorrectness is caused by the introduced non-parametric knowledge, $y_{n+p}$ can be approximately regarded as a non-parametric answer $y_n$.
It helps the model to reconsider whether to utilize the non-parametric knowledge before generation.
Ultimately, combine both subsets and obtain the training set, $\mathbf{D}_{train} = \mathbf{D}^1 \cup \mathbf{D}^2$, which consists of samples that involve knowledge conflict.

\paragraph{Supervised Fine-Tuning}
\label{sec-SFT}
In this stage, we perform SFT utilizing the instances that are collected with the methods in Section~\ref{sec-data_collection}, obtaining the subset $\mathbf{D}_{\text{SFT}}$.
Despite preference pairs are not required in the SFT stage, the subset is constructed only to collect knowledge conflict.
Since only one between parametric and non-parametric sources of the instances in $\mathbf{D}_{\text{SFT}}$ contains the correct knowledge, the model must determine which knowledge to rely on.
Therefore, SFT helps the model to preliminarily raise awareness of evaluating the quality of retrieval to support its decision.

\paragraph{Retrieval Preference Optimization}
As the previous illustration reveals, LLMs generally exhibit confusion and hallucination when accessing a context that contains different information than parametric knowledge.
To address this issue, we propose the Retrieval Preference Optimization (\MODELNAME{}) training strategy, enhancing the awareness of LLMs to focus on the retrieved context during response generation.
In detail, similar data filtering processing illustrated in Section~\ref{sec-SFT} is adopted to the dataset again with the fine-tuned policy $\pi_{\text{SFT}}$.
Meanwhile, which of the answers within the $(y_p,y_{n+p})$ pairs will be preferred is annotated by their accuracy.
The selected dataset through the SFT policy utilized for subsequent training is denoted as $\mathbf{D}_{\text{\MODELNAME{}}}$
Eventually, we conduct the \MODELNAME{} strategy by reducing the loss demonstrated in Equ.~(\ref{equ-RaPO_Loss}).
In this approach, we obtain the ultimate policy denoted as $\pi_{\text{\MODELNAME{}}}$, which implicitly conducts an integrated evaluation on retrieval within the generation.

\section{Experiments}
\begin{table*}[t]
\newcommand{\deepgrey}[1]{\textcolor[RGB]{128,128,128}{\textbf{#1}}}
  \centering
  \caption{Overall evaluation results on the test sets of four datasets. 
  Results are separated based on the generation LLMs.  
  The Column Adaptive Category indicates the category of the method if it belongs to adaptive RAG.
  \# API/LM Calls means the number of times that an API or an LM is called during an inference.
  \textbf{Bold} numbers indicate the best performance among all methods and LLMs.
  \dag indicates that due to the cost, only a part of the test set is evaluated.
  * indicates the results that are directly cited from the papers, otherwise results are reproduced by us with the consistent retrieval results.
  }
  \resizebox{1\linewidth}{!}{
  \begin{tabular}{lrccccc}
  \toprule
            &  Adaptive  &                      &  PopQA     &    NQ     &    TriviaQA   &    RGB      \\ 
  Method    &  Category  &  \#API/LM calls   &  (Accuracy) & (Accuracy)  & (Accuracy) &  (Accuracy) \\
  \midrule
  \multicolumn{7}{c}{\emph{Others}} \\
  \midrule
  RAG{\tiny ChatGPT}\dag    & -  & 1 & 50.8 & 41.8 & 65.7 & 99.3 \\
  AstuteRAG     & Pre-Eval  & 2-4    & 42.1 & 51.5 & 47.6 & 94.6 \\
  \midrule
  \multicolumn{7}{c}{\emph{LLaMA2-7B}} \\
  \midrule
  RAG           & -         & 1     & 48.8 & 22.0 & 52.5 & 91.6 \\
  RAG + SFT          & -         & 1     & 51.3 & 36.0 & 54.3 & 94.6 \\
  RAG + DPO          & -         & 1     & 53.6 & 43.5 & 51.7 & 96.3 \\
  CRAG\dag      & Pre-Eval  & 6     & 54.9 & 38.4 & 59.6 & 92.0 \\
  Self-RAG      & Post-Eval & 2-11  & 54.9 & 42.4 & \deepgrey{68.9} & 92.6 \\
  \MODELNAME{}  & Integrated-Eval & 1     & \deepgrey{55.8} & \deepgrey{45.3} & 57.6 &  \deepgrey{97.3} \\
  \midrule
  \multicolumn{7}{c}{\emph{LLaMA3-8B-instruct}} \\
  \midrule
  RAG           & -         & 1     & 59.0 & 41.3 & 65.8 & 96.3 \\
  InstructRAG   & -         & 1     & 65.0 & 46.7 & 65.1 & 99.3 \\
  Self-RAG*     & Post-Eval & 2-11  & 55.8 & 42.8 & 71.4 & - \\
  \MODELNAME{}  & Integrated-Eval & 1     & \textbf{65.4} & \textbf{51.9} & \textbf{74.4} & \textbf{100.0}\\
  \bottomrule
  \end{tabular}  
  }
  \label{tab-result}
\end{table*}

We conducted experiments to extensively demonstrate \MODELNAME{}'s advancement and adaptability to RAG-based approaches and their generalizability across various tasks.

\subsection{Tasks, Datasets and Metrics}

\MODELNAME{} was evaluated on four datasets, including \textbf{PopQA}~\cite{popqa}, \textbf{NQ}~\cite{nq}, \textbf{RGB}~\cite{rgb}, and \textbf{TriviaQA}~\cite{triviaqa}.
Following previous work, accuracy was adopted as the evaluation metric for the benchmarks.
On the one hand, the same metrics are used because our proposed method is comparable to previous studies.
On the other hand, the accuracy metric objectively measures the accuracy of the knowledge within generated responses, which appropriately represents the performance of methods in knowledge-intensive tasks.

\subsection{Baselines}

We primarily compared \MODELNAME{} with previous RAG-based baselines, which can be divided into three categories according to the base model, including:

\textbf{LLaMA2-7B} approaches utilized the vanilla or instruction-tuned LLaMA2-7B model for response generation.
(1) RAG + SFT directly tuned the model with the instances that involve knowledge conflict.
(2) RAG + DPO tuned the model with SFT in phase 1, while tuning the model with DPO rather than \MODELNAME{} in Phase 2.
Conflict collection is implemented in both SFT and DPO before training to ensure comparability.
(3) Self-RAG~\citep{self-rag} that tuned the LLaMA2 on the instruction-tuning data containing several sets of reflection tokens which were labeled by GPT-4 \citep{gpt4}, while
(4) CRAG~\citep{crag} that evaluated the quality of the retrieval and selectively corrected the retrieved context with the web search.

\textbf{LLaMA3-8B-Instruct} approaches generated the response with LLaMA3-8B-Instruct.
(1) InstructRAG~\citep{instructrag} proposes a instruction-tuning method, while
(2) Self-RAG are along with the methods above except the base model. 
Notably, results on Self-RAG with * indicate that the results are directly cited from the previous paper.

\textbf{Commercial APIs} refers to the approaches that import commercial LLMs for text generation.
We introduce the methods driven by commercial APIs for reference to benchmark the broader effectiveness and efficiency of our proposed \MODELNAME{}.
Specifically, AstuteRAG~\citep{astuterag} was reproduced in this experiment on ChatGPT, which iteratively filtered and revised the knowledge before generation.

\subsection{Results}
Table~\ref{tab-result} presents the results on four datasets. 
We briefly mark the categories of the listed adaptive RAG methods in the table.
To showcase the efficiency of \MODELNAME{} in computational overhead during the inference phase, the estimated API call or LLM inference times are presented as well.
From these results, we can conclude the following findings:

\emph{First, the proposed method significantly outperformed previous baselines that involve adaptive retrieval, reaching state-of-the-art.}
Specifically, as shown in table~\ref{tab-result}, \MODELNAME{} outperformed RAG by margins of 6.4\% accuracy on PopQA, 10.6\% accuracy on NQ, 8.6\% accuracy on TriviaQA, and 3.7\% accuracy on RGB when based on \emph{LLaMA3-8B-instruct},
as well as by margins of 7.0\% accuracy on PopQA, 23.3\% accuracy on NQ, 5.1\% accuracy on TriviaQA, and 5.7\% on RGB when based on \emph{LLaMA2-hf-7b}.
Compared with the currently advanced adaptive RAG methods, \MODELNAME{} has generally outperformed in all the benchmarks.
The advancements in our method greatly illustrate the effectiveness of preference optimization, showing the significance of overcoming the knowledge conflict.

\emph{Second, the proposed method exhibited greater computational efficiency, providing a practical solution in the real-world application for knowledge conflict mitigating.}
It can be seen that either pre-eval or post-eval approaches require multiple calls of API or LMs within a single inference.
Compared to the previous adaptive RAG, the retrieval evaluation is performed synchronously through generation.
Meanwhile, even better results are obtained, further illustrating the efficacy of our \MODELNAME{}.

\subsection{Ablation Study}
\begin{table}[t]
  \newcommand{\deepgrey}[1]{\textcolor[RGB]{128,128,128}{\textbf{#1}}}
  \centering
  \caption{Ablation study for removing retrieval-awareness, preference optimization, and SFT phases respectively on the PopQA dataset in terms of accuracy. {\small \~ w/o RR} means that the retrieval reward term is removed for optimization, while {\small \~ w/o PO} means that the model is trained without preference optimization.
  }
  \setlength{\tabcolsep}{2.0pt}
  \resizebox{0.98\linewidth}{!}{
  \begin{tabular}{lcccc}
  \toprule
                             &  {PopQA}  & NQ & TriviaQA & RGB \\ 
  \midrule
  LLaMA2-7B-hf              &  & &  &  \\
  \hdashline 
   \MODELNAME{}                         & 55.8 & 45.3 & 57.6 &  97.3 \\
   \MODELNAME{} {\small \~ w/o RR}      & 53.6 & 43.5 & 51.7 &  96.3 \\
   \MODELNAME{} {\small \~ w/o PO}      & 51.3 & 36.0 & 54.3 &  94.6 \\
   \MODELNAME{} {\small \~ w/o SFT}     & 52.5 & 34.9 & 50.1 &  90.6 \\
  \bottomrule
  \end{tabular}  
  }
  \label{tab-ablation}
\end{table}
Given that our training pipeline incorporates two distinct phases—supervised fine-tuning and preference optimization and both of which contribute to enhancing retrieval awareness and mitigating knowledge conflict, we conduct ablation studies to evaluate the individual contribution of each phase within our \MODELNAME{} framework.
The fine-tuning and preference optimization phases are removed specifically in the experiment and the results are evaluated on the benchmarks.
It is worth noting that, since the retrieval reward term in Equ.~(\ref{equ-RaPO_Loss}) is the biggest difference between DPO and \MODELNAME{}, \MODELNAME{} without the retrieval reward term (RR) can be equivalent to a DPO model.
Similarly, \MODELNAME{} without preference optimization represents models trained solely via supervised fine-tuning, omitting the subsequent alignment stage.
Results in Table~\ref{tab-ablation} demonstrate that the performance dropped when removing either phases, revealing the significance.

\subsection{Robustness to Low-Quality Retrieval}
\begin{table}[t]
  \newcommand{\deepgrey}[1]{\textcolor[RGB]{128,128,128}{\textbf{#1}}}
  \centering
  \caption{The robustness of each training strategy to low-quality retrieval in the PopQA dataset, where all retrieval information is incorrect.
  }
  \setlength{\tabcolsep}{2.0pt}
  \resizebox{0.98\linewidth}{!}{
  \begin{tabular}{ll}
  \toprule
                             &  Acc in Low-Quality Retrieval  \\ 
  \midrule
  LLaMA2-7B-hf              &  \\
  \hdashline 
   RAG                                        & ~~~~~~~~~~18.6 (0.0\%) \\
   SFT       & ~~~~~~~~~~19.5 (+4.8\%) \\
   DPO     & ~~~~~~~~~~19.3 (+3.7\%) \\
   \MODELNAME{}                               & ~~~~~~~~~~\textbf{23.5 (+26.3\%)} \\
  \bottomrule
  \end{tabular}  
  }
  \label{tab-low_quality}
\end{table}
As illustrated above, one of the primary objectives in this paper is to improve the ability of LLMs to select accurate information amidst knowledge conflicts.
It frequently occurs in a low-quality retrieval environment, posing significant challenges for prior methods.
Therefore, to further evaluate the robustness of \MODELNAME{} to low-quality retrieval, we simulate this environment by assessing the performance of LLMs when provided \emph{only} with incorrect information.
Results in Table~\ref{tab-low_quality} reveal the performance degradation of various methods under the condition of erroneous retrieval context in PopQA.
Although all methods inevitably suffer from performance degradation, our \MODELNAME{} still maintains a superior performance.
The experiments further demonstrate that unlike DPO, which exhibits limitations and potential biases when applied to RAG, \MODELNAME{} can effectively evaluate the correctness of the retrieved context during the response generation.

\subsection{Impact of Training Set Filtering}
\begin{table}[t]
  \newcommand{\deepgrey}[1]{\textcolor[RGB]{128,128,128}{\textbf{#1}}}
  \centering
  \caption{Comparison results between \MODELNAME{} with and without data filtering during SFT phase.
  }
  \setlength{\tabcolsep}{2.0pt}
  \resizebox{0.98\linewidth}{!}{
  \begin{tabular}{lcccc}
  \toprule
                             &  {PopQA}  & NQ & TriviaQA & RGB \\ 
                            
  \midrule
   LLaMA2-7B-hf                       & 48.8 & 22.0 & 52.5 & 91.6 \\
   $\pi_{\text{SFT}}$ w/o filtering   & 46.9 & 38.2 & 48.8 & 80.0 \\
   $\pi_{\text{SFT}}$ with filtering  & 51.3 & 36.0 & 54.3 & 94.6 \\
  \bottomrule
  \end{tabular}  
  }
  \label{tab-filter}
\end{table}
In phase 1 of the training stage, supervised fine-tuning is introduced for the preliminary training.
Notably, the training set is filtered, only the instances that involve knowledge conflict are selected for supervised fine-tuning.
We hypothesize that LLMs possess the inherent ability to assess retrieval quality while generating responses, albeit not activated yet.
Therefore, the operation is solely intended to enhance the retrieval awareness of LLMs, rather than to learn more knowledge.
In fact, the experimental results in table~\ref{tab-filter} reveal that the fine-tuned LLM without data filtering significantly underperformed, even worse than the original LLM before tuning, further verifying our hypothesis.

\subsection{Knowledge Selection Performance}

\begin{figure}[t]
  \centering
  \includegraphics[width=0.4\textwidth]{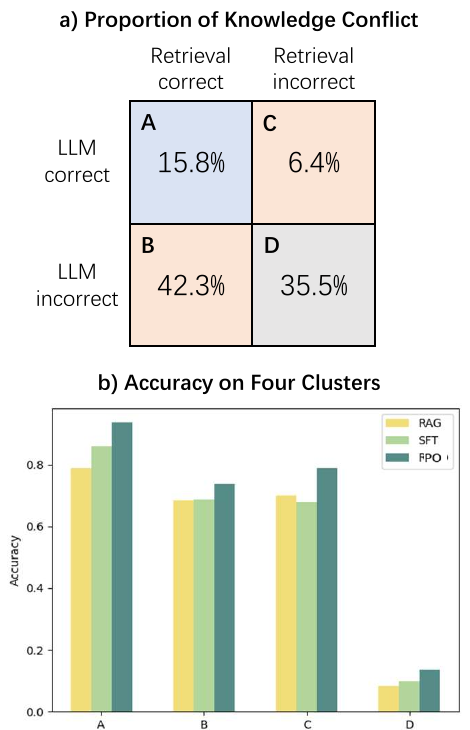}
  \caption{
  Proportion of four clusters in PopQA and the corresponding accuracy scores on LLaMA2-7B.
  }
  \label{fig-conflict}
\end{figure}
In this section, we compare \MODELNAME{} with previous training strategies in terms of knowledge selection performance. 
Further analysis is conducted on the issue of knowledge conflict before and after \MODELNAME{}.
The results in figure~\ref{fig-conflict} reveal a consistent advancement in all clusters to evaluate the knowledge and select the correct autonomously.
Besides, we found that the ability of the LLM to select knowledge can be even worse after SFT.
In Cluster B and C, which involve knowledge conflicts, SFT does not achieve a positive advancement, while \MODELNAME{} has shown a significant improvement in knowledge selection.

\section{Conclusion}

This paper studies the issue of knowledge conflict where parametric knowledge and retrieved non-parametric knowledge in RAG are inconsistent.
Previous model alignment methods have been proved limited in the context of RAG application, leading to inadequacy and bias when knowledge conflict is involved.
Therefore, a new proximal policy optimization algorithm named Retrieval Preference Optimization is proposed to adapt the RAG application.
The capability of LLMs to evaluate of the retrieval is integrated into the generation with our \MODELNAME{}, which greatly improves the efficacy compared with previous adaptive RAG approaches.
Experiments extensively demonstrate its advancement as well as generalizability across various benchmarks.
Future work will continually explore a more integrated and implicit approach for retrieval evaluation to further enhance the reliability and robustness of RAG.

\section*{Limitations}
While we primarily proposed to improve the RAG framework with a dedicated alignment method, whether a better reward function exists requires further study.
Although we make an effort to prevent reward hacking during the experiments, the intended objective can still not be fully fulfilled.
In addition, since the model is only trained on NQ, the training data could not cover various domains, leading to potential bias.
Future work will further explore a more flexible and robust rewarding strategy for RAG.

\bibliography{custom}

\clearpage
\appendix

\section{Detailed Poofs}
\label{sec-proofs}

\subsection{Proof for Equation~\ref{equ-DPO_for_RAG_Loss}}
\label{appendix-DPO_for_RAG_Loss}
In DPO optimization algorithm, a latent reward model $r(x,y)$ is adopted, which is consistent with RLHF.
To quantify the preferences, the Bradley-Terry model is introduced, which can be written as:
\begin{equation}
p(y_w \succ y_l | x) = \sigma (r(x,y_w) - r(x,y_l)),
\label{equ-BT_model}
\end{equation}
where $\sigma$ is the logistic function. 
Therefore, given a reward model  $r(y,x)$, the task can be defined as a binary classification problem and the negative log-likelihood loss can be:
\begin{flalign}
\begin{split}
\mathcal{L}_{R}(r, \mathcal{D}) =-\mathbb{E}&_{(x,y_w,y_l)\sim D}[ \log \\
& \sigma (r(x,y_w) - r(x,y_l)) ].
\label{equ-loss_RM_RLHF}
\end{split}
\end{flalign}

If DPO is directly adopted for RAG and taking $y_n$ and $y_p$ as the $(y_w, y_l)$ pair, considering the influence of retrieved context $D^r$, the expression of reward model would get modified as:

\begin{flalign}
 \label{equ-RaPO_RM_para} r(x,y_p) = \beta\log & {\frac{\pi_\theta(y_p|x)}{\pi_{ref}(y_p|x)}} + \beta \log{Z(x)} ; \\
\begin{split}
 \label{equ-RaPO_RM_nonpara} r(x,y_n) = \beta\log & {\frac{\pi_\theta(y_n|x,D^r)}{\pi_{ref}(y_p|x,D^r)}} \\
 &\quad\quad\quad\ + \beta \log{Z(x,D^r)} .
\end{split}
\end{flalign}

Substituting the representation in Equ.~(\ref{equ-RaPO_RM_para}) and (\ref{equ-RaPO_RM_nonpara}) for the Bradley-Terry model in Equ.~(\ref{equ-BT_model}), it can be found that the partition function can not be canceled.

\subsection{Proof for Equation~\ref{equ-sub_solution}}
\label{appendix-sub_solution}
In order to apply DPO to RAG, fabricated answers are necessary.
However, the fabricated answer is may not the candidate answers with the highest likelihood for LLMs, i.e., existing $y_p$, satisfying that:
\begin{equation}
\begin{cases}
\log\pi_{\text{ref}}(y_n\mid x, D^r)> \epsilon\\
\log\pi_{\text{ref}}(y_p\mid x, D^r)< \epsilon\\
\log\pi_{\text{ref}}(y_n\mid x, D^r)-\log\pi_{\text{ref}}(y_p\mid x, D^r)>\epsilon_{d},
\label{equ-preliminary_fab_ans}
\end{cases}
\end{equation}
where $\epsilon_{d}$ indicates the difference of logits between parametric output and non-parametric output, which can be massive. While $\pi_{\text{SFT}}$ is $\pi_{\text{ref}}$, which is used as the reference policy in the optimization phase.

It could lead to a concern that the optimized LLMs would not converge to the optimal solution.
Two aspects can theoretically interpret the conclusion.
On the one hand, the proposal of the DPO reward model training strategy comes from the RL optimization objective of RLHF, as shown in Equ.~(\ref{equ-RLHF_RL_obj}).
Therefore, due to the constraint of the KL-divergence, the distribution of the policy would not change a lot, i.e.:
\begin{equation}
\begin{cases}
\left | \log\pi_{\theta}(y_n\mid x, D^r)-\log\pi_{\text{SFT}}(y_n\mid x, D^r) \right |< \epsilon_{ad}\\
\left | \log\pi_{\theta}(y_p\mid x, D^r)-\log\pi_{\text{SFT}}(y_p\mid x, D^r) \right |< \epsilon_{ad},
\label{equ-preliminary_fab_ans_1}
\end{cases}
\end{equation}
where $\epsilon_{ad} > 0$ is a very limited value.
Supposing a situation during the inference $(x_{\text{inf}}, y_{p\_\text{inf}}\succ y_{n\_\text{inf}})$ that can generally exist, where the parametric answer is wining, meanwhile, the distance between parametric and non-parametric is big enough so that $\epsilon_{d} > 2\epsilon_{ad}$, then the generator would still choose the losing one as the ultimate response:
\begin{flalign}
\begin{split}
\pi_{\theta}(y_w\mid x_{\text{inf}}, D^r) &= \pi_{\theta}(y_{p\_\text{inf}}\mid x, D^r)\\
&<\pi_{\text{ref}}(y_{p\_\text{inf}}\mid x_{\text{inf}}, D^r)+\epsilon_{ad}\\
&<\pi_{\text{ref}}(y_{n\_\text{inf}}\mid x_{\text{inf}}, D^r)-\epsilon_{ad}\\
&<\pi_{\theta}(y_{n\_\text{inf}}\mid x_{\text{inf}}, D^r)\\
&=\pi_{\theta}(y_l\mid x_{\text{inf}}, D^r).
\nonumber
\end{split}
\end{flalign}

\subsection{Derivation of \MODELNAME{}'s Reward Model}
\label{appendix-rapo_rm}
Given the RL objective as Equ.~(\ref{equ-RaPO_RL_obj}) shows, expanding the KL-divergence Formula and derive:
\begin{flalign}
\begin{split}
\max_{\pi_\theta}&{\mathbb{E}}r (x,y,R)\\
&-  \beta\mathbb{D}_{\text{KL}}[\pi _\theta (y,R\mid x) \left |  \right | \pi _{\text{ref}}(y,R\mid x) ] .\\
= \min_{\pi_\theta}&{\mathbb{E}}\left [ \log{\frac{\pi_{\theta}(y,R\mid x)}{\pi_{\text{ref}}(y,R\mid x)}-\frac{1}{\beta} r(x,y,R)} \right ],
\end{split}
\end{flalign}
while following the previous work\citep{DBLP:conf/icml/PetersS07,DBLP:journals/corr/abs-1910-00177,DBLP:conf/nips/KorbakEKD22,DBLP:conf/icml/GoKKRRD23,dpo}, it is straightforward to show that the optimal solution takes the form:
\begin{flalign}
\begin{split}
\pi_{r}(y,R\mid x)=\frac{\pi_{\text{ref}}(y,R\mid x)\exp{(\frac{1}{\beta}r(x,y,R))}}{Y(x)},
\label{equ-rm_equation}
\end{split}
\end{flalign}
where the partition function can be formulated as:
\begin{flalign}
\begin{split}
Y(x) =& \sum_{y}\sum_{R}\pi_{\text{ref}}(y,R\mid x)\exp{(\frac{1}{\beta}r(x,y,R))}\nonumber.
\end{split}
\end{flalign}

Based on Equ.~(\ref{equ-rm_equation}), the reward model can be derived and written as:
\begin{flalign}
\begin{split}
r(x,y,R) =&\beta\log{\frac{\pi(y,R\mid x)}{\pi_{\text{ref}}(y,R\mid x)}}
+ \beta\log{Y(x)}.
\end{split}
\end{flalign}
Following the Bayes theorem, the reward model can be formulated as Equ.~(\ref{equ-rm}).

\section{Experiment Details}

\subsection{Details of the Datasets}
\MODELNAME{} was evaluated on four datasets, which are in public domain and licensed for research purposes, including:

\textbf{PopQA}~\cite{popqa} is a \emph{short}-form generation task. 
Generally, only one entity of factual knowledge is expected to be answered for each single question. 
In our experiments, we exactly followed the setting in the previous work~\cite{self-rag} which evaluated methods on a long-tail subset consisting of 1,399 rare entity queries whose monthly Wikipedia page views are less than 100.

\textbf{Natural Questions} (\textbf{NQ})~\cite{nq} is a benchmark for question answering research that contains real user questions issued to Google search, and answers found from Wikipedia by annotators. 
Annotations include long answers (usually a paragraph of text) and short answers (one or more entities), which are marked as null if there is no answer on the page.
Additionally, NQ contains 307,372 training examples, 7,830 examples for development, and we withold a further 7,842 examples for testing.
Only short answers are adopted in our experiments.

\textbf{TriviaQA}~\cite{triviaqa} is a reading comprehension dataset containing over 650K question-answer-evidence triples. TriviaQA includes 95K question-answer pairs authored by trivia enthusiasts and independently gathered evidence documents, six per question on average, that provide high quality distant supervision for answering the questions.

\textbf{Retrival-Augmented Generation Benchmark} (\textbf{RGB})~\cite{rgb} is a benchmark that chooses to aggregate the latest news.
Different basic abilities of LLMs are evaluated according to the common challenges in RAG, including noise robustness, negative rejection, information integration and counterfactual robustness.

\subsection{Experimental Setup}
We use the package \emph{vllm} for inference, and the parameter settings are listed below:
{
\texttt{\\
temperature=0.0\\
top\_p=1.0\\
max\_tokens=100\\
skip\_special\_tokens=false\\
}
}

The model was trained on 4*A100 in our experiment, and the SFT was implemented with the hyperparameter settings below:
{
\texttt{\\
n\_epochs=1\\
batch\_size=4\\
gradient\_accumulation\_steps=32\\
mixed\_precision=bf16\\
max\_seq\_length=2048\\
warmup\_ratio=0.03\\
learning\_rate=2e-5\\
weight\_decay=0.0,\\
}
}
while RPO strictly followed the hyperparameters used in~\citet{dpo}.



\end{document}